\documentclass[final]{cvpr}
\usepackage{subfigure}
\usepackage{times}
\usepackage{epsfig}
\usepackage{graphicx}
\usepackage{amsmath}
\usepackage{amssymb}
\usepackage{multirow}
\usepackage{comment}
\usepackage{appendix}

\usepackage[pagebackref=true,breaklinks=true,colorlinks,bookmarks=false]{hyperref}

\begin{document}

\title{1st Place Solution to the EPIC-Kitchens Action Anticipation Challenge 2022}

\author{Zeyu Jiang$^1$
 \quad Changxing Ding$^1$\thanks{Corresponding author. } \\
$^1$ South China University of Technology  \\
{\tt\small jzy\_scut@outlook.com, chxding@scut.edu.cn}
}

\maketitle


\begin{abstract}
In this report, we describe the technical details of our submission to the EPIC-Kitchens Action Anticipation Challenge 2022. In this competition, we develop the following two approaches. 1) Anticipation Time Knowledge Distillation using the soft labels learned by the teacher model as knowledge to guide the student network to learn the information of anticipation time; 2) Verb-Noun Relation Module for building the relationship between verbs and nouns. Our method achieves state-of-the-art results on the testing set of EPIC-Kitchens Action Anticipation Challenge 2022.
\end{abstract}

\section{Introduction}
\label{sec:introduction}
EPIC-KITCHENS is a large annotated egocentric dataset \cite{damen2018scaling,damen2022rescaling}.  Action anticipation is an important task in EPIC-KITCHENS.

We summarize our main contributions as follows:

1) Aiming at the problem that the missing information of anticipation time affects the performance of egocentric action anticipation, we propose Anticipation Time Knowledge Distillation(ATKD) to distill the information of anticipation time.

2) Because of the lack of consideration of the relationship between verbs and nouns in the existing research work on Egocentric Action Anticipation, we propose a Verb-Noun Relation Module(VNRM) to model the relationship between verbs and nouns.

3) Our approaches show superior results on EPIC-KITCHENS-100.

\begin{figure}[t]
	\begin{center}
		\includegraphics[width=0.45\textwidth]{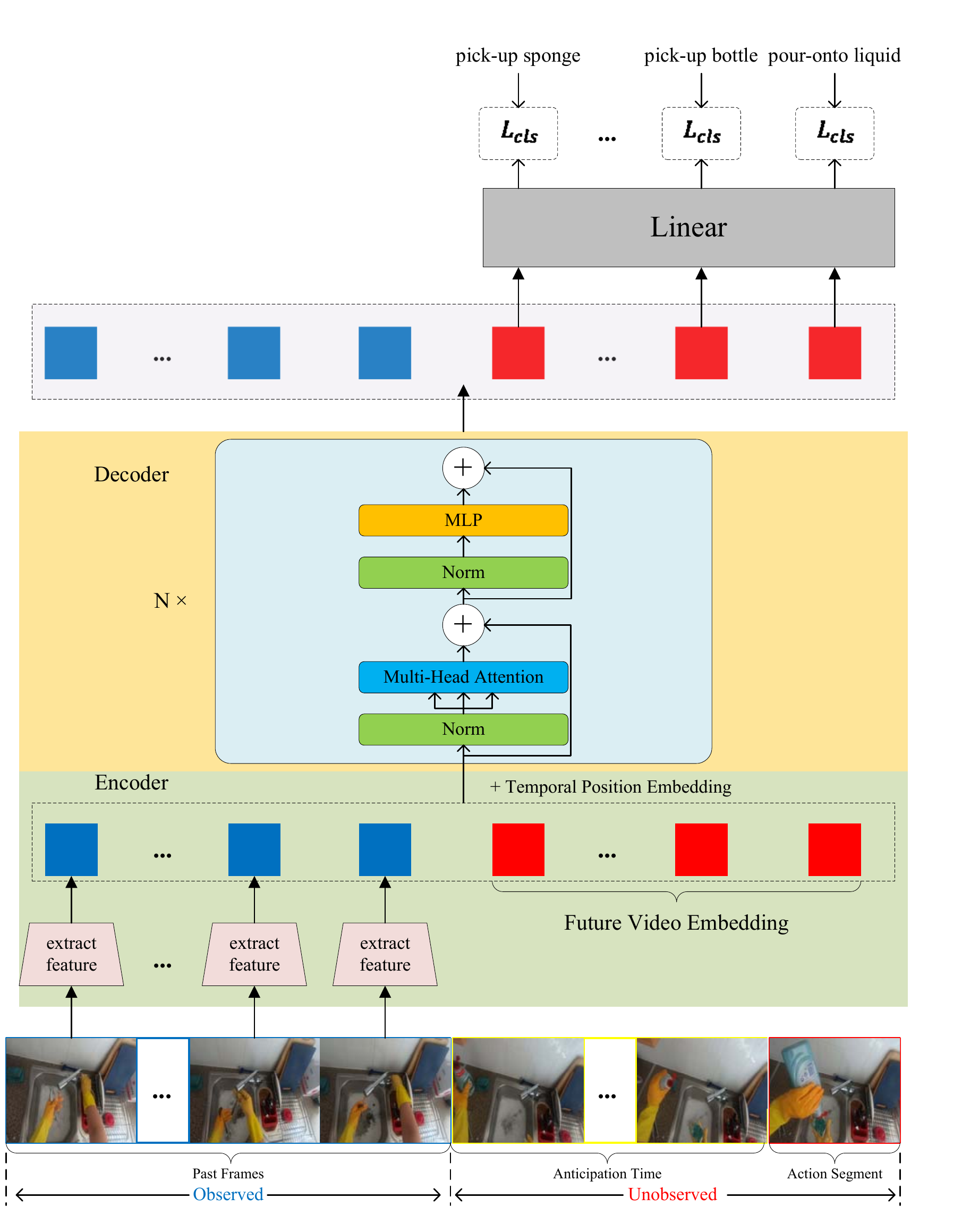}
	\end{center}
	\caption{ Student model.}
	\label{Figure:student}
\end{figure}

\section{Our approach}
\label{sec:method}
\subsection{Base Model}
We use Causal Transformer Decoder (like AVT-h)\cite{girdhar2021anticipative} as base model. We use a 4-head, 4-layer model as our baseline.

\subsection{Anticipation Time Knowledge Distillation(ATKD)}
\begin{figure}[t]
	\begin{center}
		\includegraphics[width=0.45\textwidth]{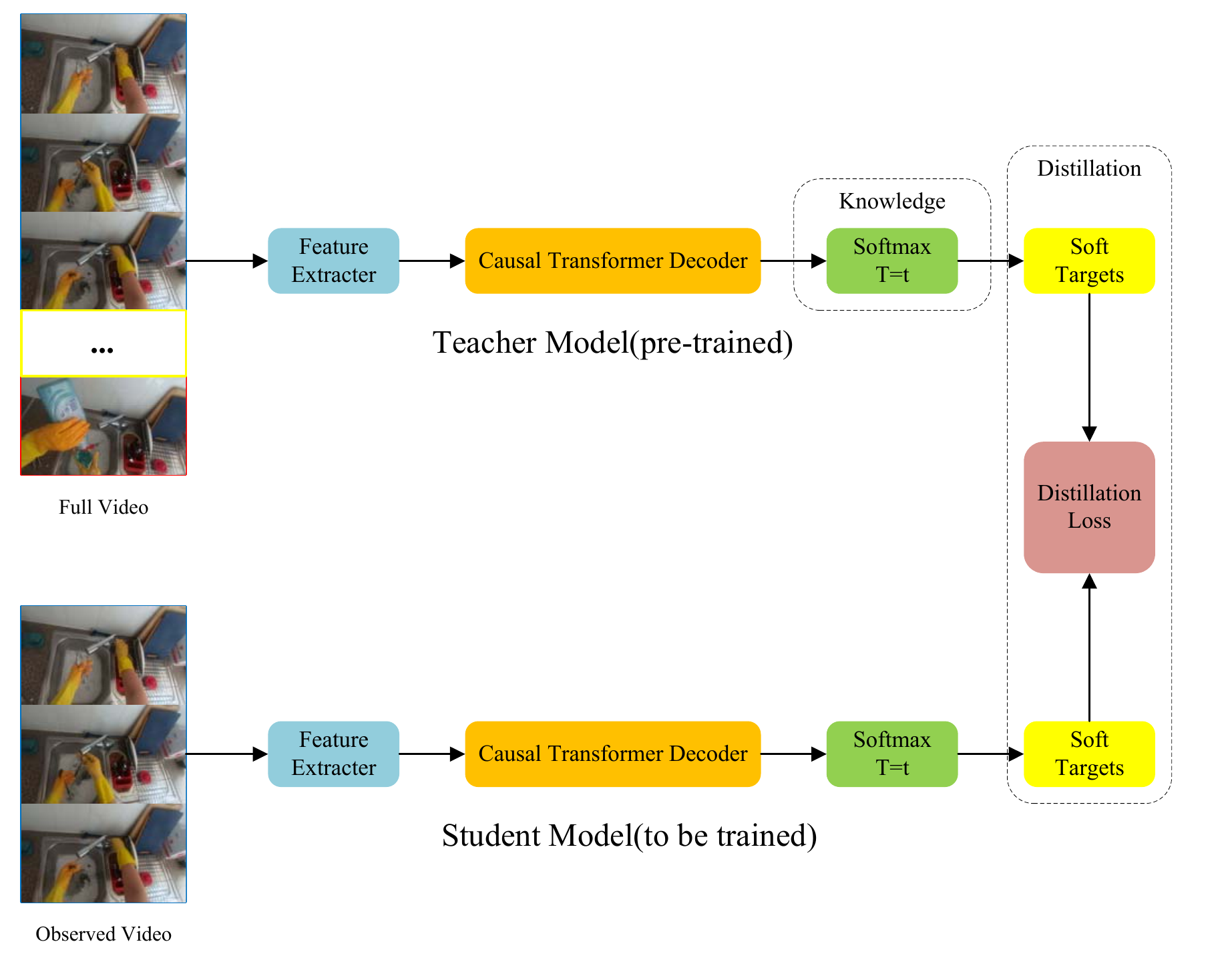}
	\end{center}
	\caption{ Overview of Anticipation Time Knowledge Distillation.}
	\label{Figure:atkd}
\end{figure}

The temporal gap between the past observations and the future action(Anticipation Time)\cite{furnari2019would,wu2020learning} will result in missing information. To solve the problem that the missing information of anticipation time, we propose a knowledge distillation method to distill the information of anticipation time. Fig.~\ref{Figure:student} shows the student model. We initialize the future video embedding with the learnable parameter. Fig.~\ref{Figure:atkd} shows the overview of Anticipation Time Knowledge Distillation. The input of the teacher model is full video and the input of the student model is the concatenation of the observed video and future video embedding. In the teacher model, if there are no labels in the anticipation time clip, we use the label of the closest labeled clip as its label. The teacher model can distill the soft label of anticipation time to the student model.

\begin{figure}[h]
	\begin{center}
		\includegraphics[width=0.45\textwidth]{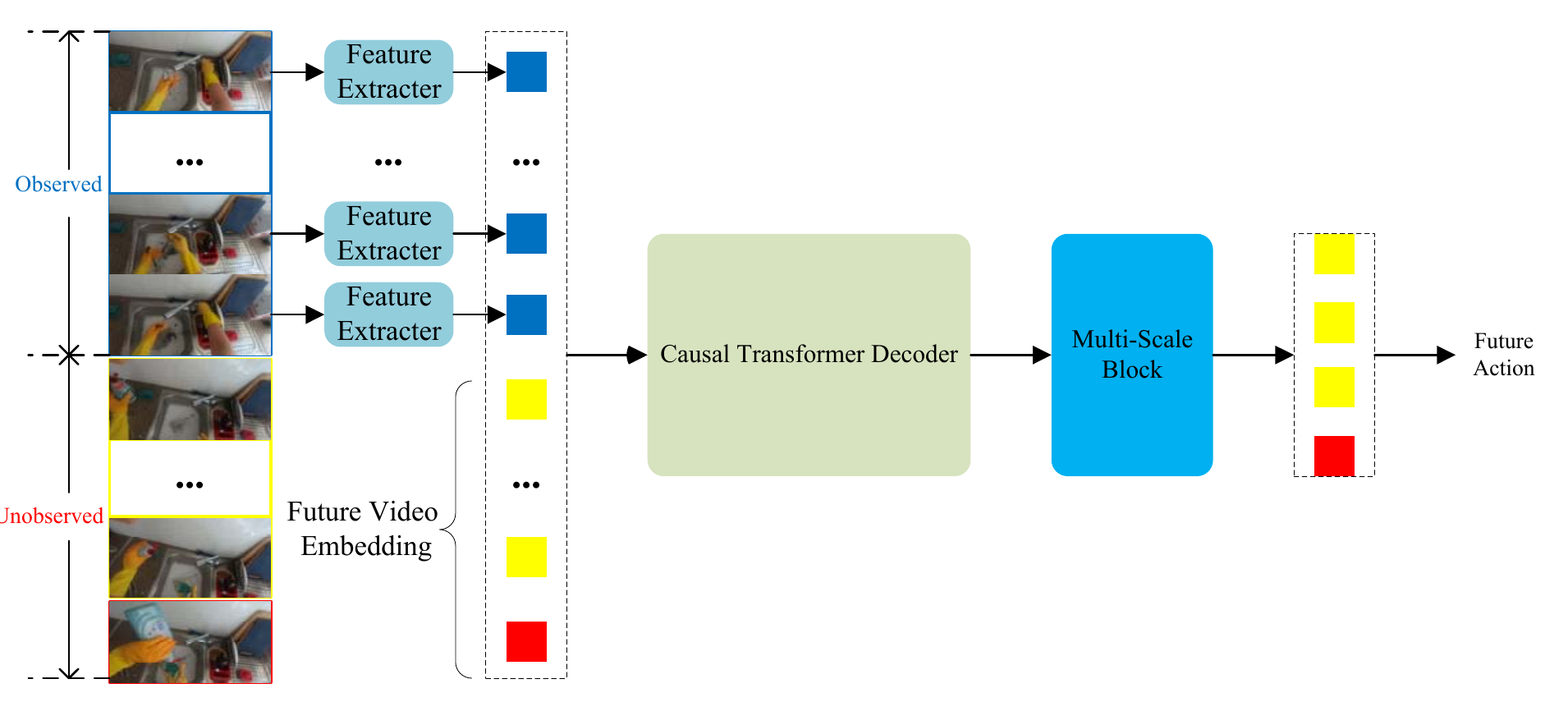}
	\end{center}
	\caption{ Student model with multi-scale block.}
	\label{Figure:mss}
\end{figure}

\begin{figure}[h]
	\begin{center}
		\includegraphics[width=0.45\textwidth]{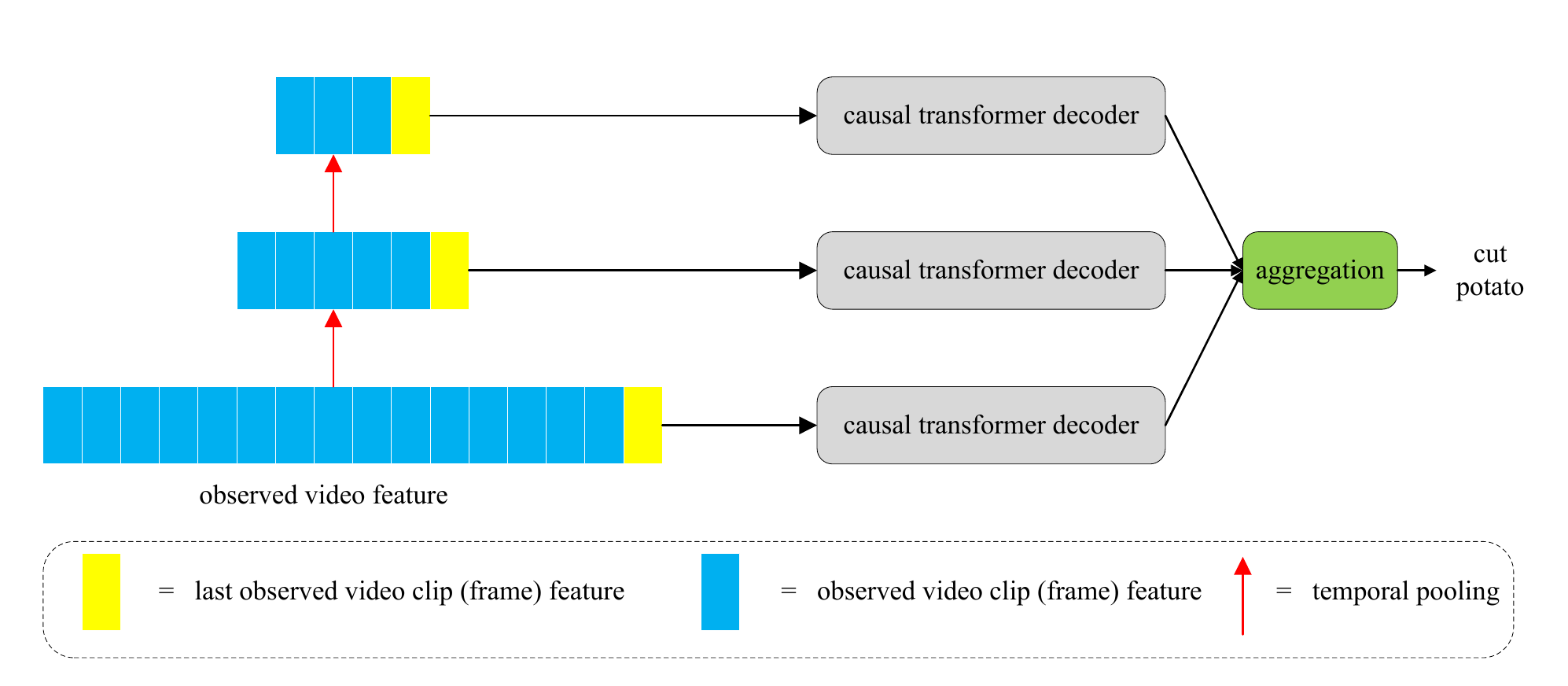}
	\end{center}
	\caption{ Multi-scale block.}
	\label{Figure:ms}
\end{figure}

Finally, we use a multi-scale block to improve the performance. Fig.~\ref{Figure:mss} shows the architecture of  student model with multi-scale block. Fig.~\ref{Figure:ms} shows the details of multi-scale block.

\subsection{ Verb-Noun Relation Module(VNRM)}
\begin{figure}[t]
	\begin{center}
		\includegraphics[width=0.45\textwidth]{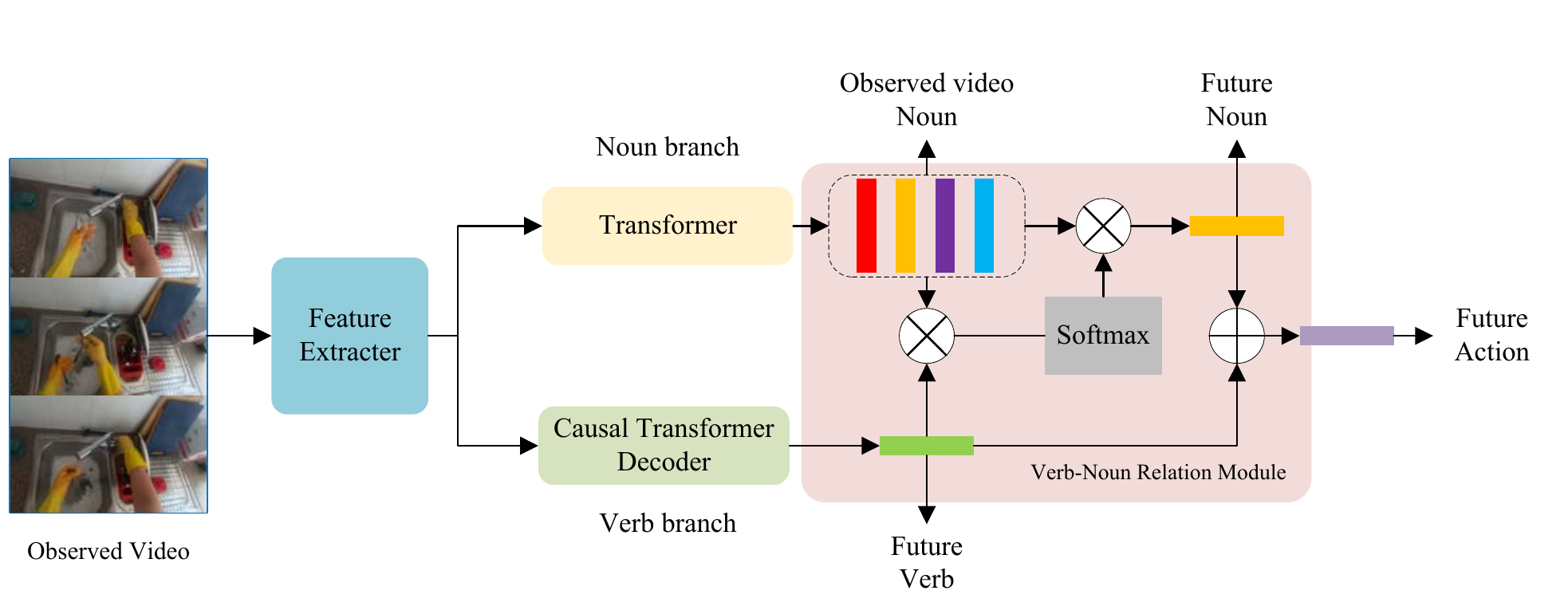}
	\end{center}
	\caption{ Overview of Verb-Noun Relation Module.}
	\label{Figure:vnrm}
\end{figure}

Inspired by \cite{wang2020symbiotic} and \cite{wang2021interactive}, we propose a verb-noun relationship interaction module to model the relationship between verbs and nouns. The module guides the features of the nouns interacting with the wearer in the observed videos to represent the features of the nouns interacting with the wearer in the future through the features of the predicted future verbs. Fig.~\ref{Figure:vnrm} shows the overview of Verb-Noun Relation Module.

The same as Anticipation Time Knowledge Distillation, if there are no labels in the clip, we use the label of the closest labeled clip as its label.

Finally, we use knowledge distillation to improve the performance. The input of the teacher model's verb branch is the full video and the input of the teacher model's noun branch is only the observed video.

\subsection{Feature Extraction}

We use some action recognition models as backbones to extract features.

The backbones are as follow:

\noindent{\bf Model A}
SlowFast 16$\times$8, R101$+$NL\cite{feichtenhofer2019slowfast}, predicting verb and noun

\noindent{\bf Model B}
SlowFast 8$\times$8, R101\cite{feichtenhofer2019slowfast}, predicting verb and noun

\noindent{\bf Model C}
TSN(BNInception)\cite{furnari2019would,wang2016temporal}

\noindent{\bf Model D}
Mformer-L\cite{patrick2021keeping}, with temporal stride 4

\noindent{\bf Model E}
Mformer-HR\cite{patrick2021keeping}, with temporal stride 8

\noindent{\bf Model F}
Mformer-HR\cite{patrick2021keeping}, with temporal stride 4

\noindent{\bf Model G}
SlowFast 16$\times$8, R101$+$NL\cite{feichtenhofer2019slowfast}, predicting verb, noun and action

\noindent{\bf Model H}
SlowFast 8$\times$8, R101\cite{feichtenhofer2019slowfast}, predicting verb, noun and action

\subsection{Ensemble}
We use an ensemble of a set of 10 models as final result for testing set.

\section{Experiments}

\subsection{Implementation Details}
 We train the networks using AdamW\cite{loshchilov2017decoupled}, using a batch size of 128,  label smoothing\cite{szegedy2016rethinking} of 0.4, an l2 weight decay of $5 e-4$, and an initial learning rate of  $1 e-4$. The maximum number of training iterations is set to 300 epochs. A cosine annealing with a warm-up restart schedule (20 cycles) is used. The  cycle is set to 15 epochs with 1 epoch of linear warmup. All 10 models which we use as an ensemble for the testing set are trained on the same hyperparameters with the same random seed.

\begin{table}[htb]
\caption{Results of ablation studies(Anticipation Time Knowledge Distillation).}\label{tab0}
\resizebox{0.45\textwidth}{!}{
\begin{tabular}{llllllll}
\hline
Method     & KD & Multi-Scale & Backbone & Backbone(teacher)       & Verb & Noun & Action \\ \hline
Base Model &    &             & F        & \textbackslash{}        & 32   & 32.3 & 15.9   \\
ATKD       &    &             & F        & \textbackslash{}        & 31.2 & 34.6 & 16.7   \\
ATKD       & $\checkmark$  &             & F        & F                       & 32.2 & 35.3 & 17.3   \\
ATKD       & $\checkmark$  & $\checkmark$           & F        & F                       & 31.7 & 36.4 & 18.1   \\
ATKD       & $\checkmark$  & $\checkmark$           & F        & B                       & 33.7 & 36.3 & 19.1   \\
ATKD       & $\checkmark$  & $\checkmark$           & F        & B+F(average soft label) & 36.5 & 36.8 & 18.7   \\ \hline
\end{tabular}}
\end{table}

\begin{table}[htb]
\caption{Results of ablation studies(Verb-Noun Relation Module).}\label{tab01}
\resizebox{0.45\textwidth}{!}{
\begin{tabular}{llllll}
\hline
Method     & Backbone & Backbone(teacher)       & Verb & Noun & Action \\ \hline
Base Model & F        & \textbackslash{}        & 32   & 32.3 & 15.9   \\
VNRM(w/o KD)       & F        & \textbackslash{}        & 33.9 & 34.7 & 16.8   \\
VNRM       & F        & F                       & 31.7 & 37   & 17.5   \\
VNRM       & F        & B                       & 34.7 & 38.4 & 18.7   \\
VNRM       & F        & B+F(avreage soft label) & 32.9 & 39.7 & 19.2   \\ \hline
\end{tabular}}
\end{table}

\begin{table}[htb]
\caption{10 model for ensemble.}\label{tab1}
\resizebox{0.45\textwidth}{!}{
\begin{tabular}{cccccccc}
\hline
\#       & Method     & Backbone         & Backbone(teacher)       & Verb & Noun & Action &                                                                               \\ \hline
1        & base model & C                & \textbackslash{}        & 27.1 & 27.4 & 12.9   &                                                                               \\
2        & ATKD       & F                & B                       & 33.7 & 36.3 & 19.1   &                                                                               \\
3        & ATKD       & E                & E                       & 31.5 & 35.8 & 17.7   &                                                                               \\
4        & ATKD       & A                & A                       & 32.6 & 34.6 & 17     &                                                                               \\
5        & ATKD       & B                & B                       & 32.6 & 35.4 & 16.9   &                                                                               \\
6        & ATKD       & F                & \textbackslash{}        & 31.2 & 34.6 & 16.7   & \begin{tabular}[c]{@{}c@{}}only student model\\  w/o multi-scale\end{tabular} \\
7        & VNRM       & G                & G                       & 29.6 & 36   & 16.3   &                                                                               \\
8        & VNRM       & H                & H                       & 33.1 & 34.4 & 15.9   &                                                                               \\
9        & VNRM       & D                & B+F(average soft label) & 31.7 & 38.2 & 17.1   &                                                                               \\
10       & VNRM       & F                & B+F(average soft label) & 32.9 & 39.7 & 19.2   &                                                                               \\ \hline
ensemble & ensemble   & \textbackslash{} & \textbackslash{}        & 41   & 44.2 & 22.7   &                                                                               \\ \hline
\end{tabular}}
\end{table}

\begin{table}[htb]
\caption{The result on the testing set.(User:hrgdscs)}\label{tab2}
\resizebox{0.45\textwidth}{!}{
\begin{tabular}{cccccccccc}
\hline
         & \multicolumn{3}{c}{Overall}           & \multicolumn{3}{c}{Unseen}            & \multicolumn{3}{c}{Tail}              \\ \cline{2-10} 
         & \multicolumn{3}{c}{Mean Top-5 Recall} & \multicolumn{3}{c}{Mean Top-5 Recall} & \multicolumn{3}{c}{Mean Top-5 Recall} \\ \cline{2-10} 
         & Verb        & Noun       & Action       & Verb        & Noun        & Action      & Verb        & Noun       & Action       \\ \hline
Ensemble & 37.91       & 41.71      & 20.43      & 27.94       & 37.07       & 18.27     & 32.43       & 36.09      & 17.11      \\ \hline
\end{tabular}}
\end{table}

\subsection{Results}
The result of the ablation studies can be found in Table~\ref{tab0} and Table~\ref{tab01}. The result of 10 models for the ensemble is shown in Table~\ref{tab1}.

The final ensemble result on the testing set is presented in Table~\ref{tab2}.  Our algorithm achieved the best performance.

\section{Conclusion}
In this paper, we propose two novel methods. The validation and testing results show that our proposed method can achieve excellent performance.
{\small

\bibliographystyle{ieee_fullname}

\begin{thebibliography}{10}\itemsep=-1pt

\bibitem{damen2018scaling}
Dima Damen, Hazel Doughty, Giovanni~Maria Farinella, Sanja Fidler, Antonino
  Furnari, Evangelos Kazakos, Davide Moltisanti, Jonathan Munro, Toby Perrett,
  Will Price, et~al.
\newblock Scaling egocentric vision: The epic-kitchens dataset.
\newblock In {\em Proceedings of the European Conference on Computer Vision
  (ECCV)}, pages 720--736, 2018.

\bibitem{damen2022rescaling}
Dima Damen, Hazel Doughty, Giovanni~Maria Farinella, Antonino Furnari,
  Evangelos Kazakos, Jian Ma, Davide Moltisanti, Jonathan Munro, Toby Perrett,
  Will Price, et~al.
\newblock Rescaling egocentric vision: Collection, pipeline and challenges for
  epic-kitchens-100.
\newblock {\em International Journal of Computer Vision}, 130(1):33--55, 2022.

\bibitem{feichtenhofer2019slowfast}
Christoph Feichtenhofer, Haoqi Fan, Jitendra Malik, and Kaiming He.
\newblock Slowfast networks for video recognition.
\newblock In {\em Proceedings of the IEEE/CVF International Conference on
  Computer Vision}, pages 6202--6211, 2019.

\bibitem{furnari2019would}
Antonino Furnari and Giovanni~Maria Farinella.
\newblock What would you expect? anticipating egocentric actions with
  rolling-unrolling lstms and modality attention.
\newblock In {\em Proceedings of the IEEE/CVF International Conference on
  Computer Vision}, pages 6252--6261, 2019.

\bibitem{girdhar2021anticipative}
Rohit Girdhar and Kristen Grauman.
\newblock Anticipative video transformer.
\newblock In {\em Proceedings of the IEEE/CVF International Conference on
  Computer Vision}, pages 13505--13515, 2021.

\bibitem{loshchilov2017decoupled}
Ilya Loshchilov and Frank Hutter.
\newblock Decoupled weight decay regularization.
\newblock {\em arXiv preprint arXiv:1711.05101}, 2017.

\bibitem{patrick2021keeping}
Mandela Patrick, Dylan Campbell, Yuki Asano, Ishan Misra, Florian Metze,
  Christoph Feichtenhofer, Andrea Vedaldi, and Jo{\~a}o~F Henriques.
\newblock Keeping your eye on the ball: Trajectory attention in video
  transformers.
\newblock {\em Advances in Neural Information Processing Systems}, 34, 2021.

\bibitem{szegedy2016rethinking}
Christian Szegedy, Vincent Vanhoucke, Sergey Ioffe, Jon Shlens, and Zbigniew
  Wojna.
\newblock Rethinking the inception architecture for computer vision.
\newblock In {\em Proceedings of the IEEE conference on computer vision and
  pattern recognition}, pages 2818--2826, 2016.

\bibitem{wang2016temporal}
Limin Wang, Yuanjun Xiong, Zhe Wang, Yu Qiao, Dahua Lin, Xiaoou Tang, and
  Luc~Van Gool.
\newblock Temporal segment networks: Towards good practices for deep action
  recognition.
\newblock In {\em European conference on computer vision}, pages 20--36.
  Springer, 2016.

\bibitem{wang2020symbiotic}
Xiaohan Wang, Yu Wu, Linchao Zhu, and Yi Yang.
\newblock Symbiotic attention with privileged information for egocentric action
  recognition.
\newblock In {\em Proceedings of the AAAI Conference on Artificial
  Intelligence}, volume~34, pages 12249--12256, 2020.

\bibitem{wang2021interactive}
Xiaohan Wang, Linchao Zhu, Heng Wang, and Yi Yang.
\newblock Interactive prototype learning for egocentric action recognition.
\newblock In {\em Proceedings of the IEEE/CVF International Conference on
  Computer Vision}, pages 8168--8177, 2021.

\bibitem{wu2020learning}
Yu Wu, Linchao Zhu, Xiaohan Wang, Yi Yang, and Fei Wu.
\newblock Learning to anticipate egocentric actions by imagination.
\newblock {\em IEEE Transactions on Image Processing}, 30:1143--1152, 2020.

\end{thebibliography}

}

\end{document}